\def\year{2018}\relax
\documentclass[letterpaper]{article} 
\usepackage{aaai18}  
\usepackage{times}  
\usepackage{helvet}  
\usepackage{courier}  
\usepackage{url}  
\usepackage{graphicx}  

\usepackage{bm}
\usepackage{amsfonts}
\usepackage{mathrsfs}
\usepackage{graphicx}
\usepackage{subfigure}
\usepackage{multirow}
\usepackage{url}

\begin{document}
%
\title{RAN4IQA: Restorative Adversarial Nets\\ for No-Reference Image Quality Assessment}
\author{Hongyu Ren$^1$, Diqi Chen$^{2,1}$, Yizhou Wang$^1$\\
$^1$Nat'l Engineering Laboratory for Video Technology, Cooperative Medianet Innovation Center \\
$^1$Key Laboratory of Machine Perception (MoE), School of Electronics Engineering and Computer Science, Peking University\\
$^2$Key Laboratory of Intelligent Information Processing of Chinese Academy of
Sciences (CAS),\\ 
$^2$Institute of Computing Technology, CAS, Beijing, 100190, China\\
$^2$University of Chinese Academy of Sciences, Beijing, 100049, China\\
\{rhy, cdq, yizhou.wang\}@pku.edu.cn
}
\maketitle
\begin{abstract}
Inspired by the free-energy brain theory~\cite{friston2006free}, which implies that human visual system (HVS) tends to reduce uncertainty and restore perceptual details upon seeing a distorted image~\cite{friston2010free}, we propose restorative adversarial net (RAN), a GAN-based model for no-reference image quality assessment (NR-IQA). RAN, which mimics the process of HVS, consists of three components: a restorator, a discriminator and an evaluator. The restorator restores and reconstructs input distorted image patches, while the discriminator distinguishes the reconstructed patches from the pristine distortion-free patches. After restoration, we observe that the perceptual distance between the restored and the distorted patches is monotonic with respect to the distortion level. We further define Gain of Restoration (GoR) based on this phenomenon. The evaluator predicts perceptual score by extracting feature representations from the distorted and restored patches to measure GoR. Eventually, the quality score of an input image is estimated by weighted sum of the patch scores. Experimental results on Waterloo Exploration, LIVE and TID2013 show the effectiveness and generalization ability of RAN compared to the state-of-the-art NR-IQA models.
\end{abstract}
\section{Introduction}

\label{sec:intro}
With mobile devices ingrained in the very fabric of people's life, capture, transmission and storage of digital images are ubiquitous nowadays. People are paying closer attention to the perceptual quality of digital images. This quantitative evaluation can either be subjective or objective. However, with such a great load of images, assessing the perceptual quality merely by the human is next to impossible. Objective image quality assessment (IQA), which measures visual quality by mimicking human perception, has become an alternative solution.

In cases where the distortion-free image is available, IQA can be carried out by comparing the distorted image with the reference image. Current full-reference image quality assessment (FR-IQA) metrics such as FSIM~\cite{zhang2011fsim} and VSI~\cite{zhang2014vsi} have achieved satisfying results that are concordant with the human visual system (HVS). However, reference images seldom exist in most situations, and thus no-reference image quality assessment (NR-IQA) which evaluates perceptual quality with no access to reference images is highly demanded. Traditional learning-based IQA methods follow a routine pipeline, namely extract features first then train a model to estimate ground-truth perceptual score. 
However, none of them take into consideration the detailed process in which humans perceive and assess distorted images, which results in inconsistency with subjective evaluation.

Recent research in brain and neuroscience has brought insights into how HVS works when quantifying the quality of an image. \cite{friston2010free} propose a unified brain theory based on free-energy principle~\cite{friston2006free}, which reveals that human brain resists a natural tendency to disorder and HVS is inclined to restore and reconstruct degraded contents when assessing a distorted image. It shows that human brain will instinctively add textures and details based on the perceived image in order to simply unveil the mask of distortion and figure out the pristine content. 
After the restoration, human brain compares the distorted image to the brain-restored image in order to assess perceptual quality. Meanwhile, we also observe the correlation between distortion level and summation of brain-restored details: when people are shown images with higher level distortion, they are able to restore more details, although not all the complementary details are ground truth.

\begin{figure*}[htb]
\includegraphics[width=\textwidth]{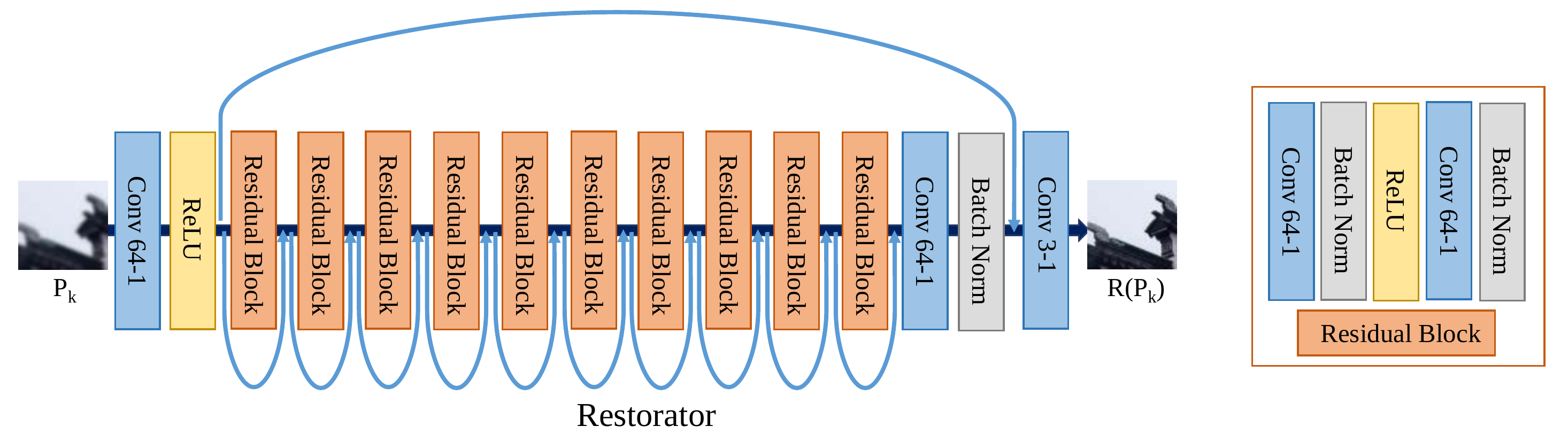}
\includegraphics[width=\textwidth]{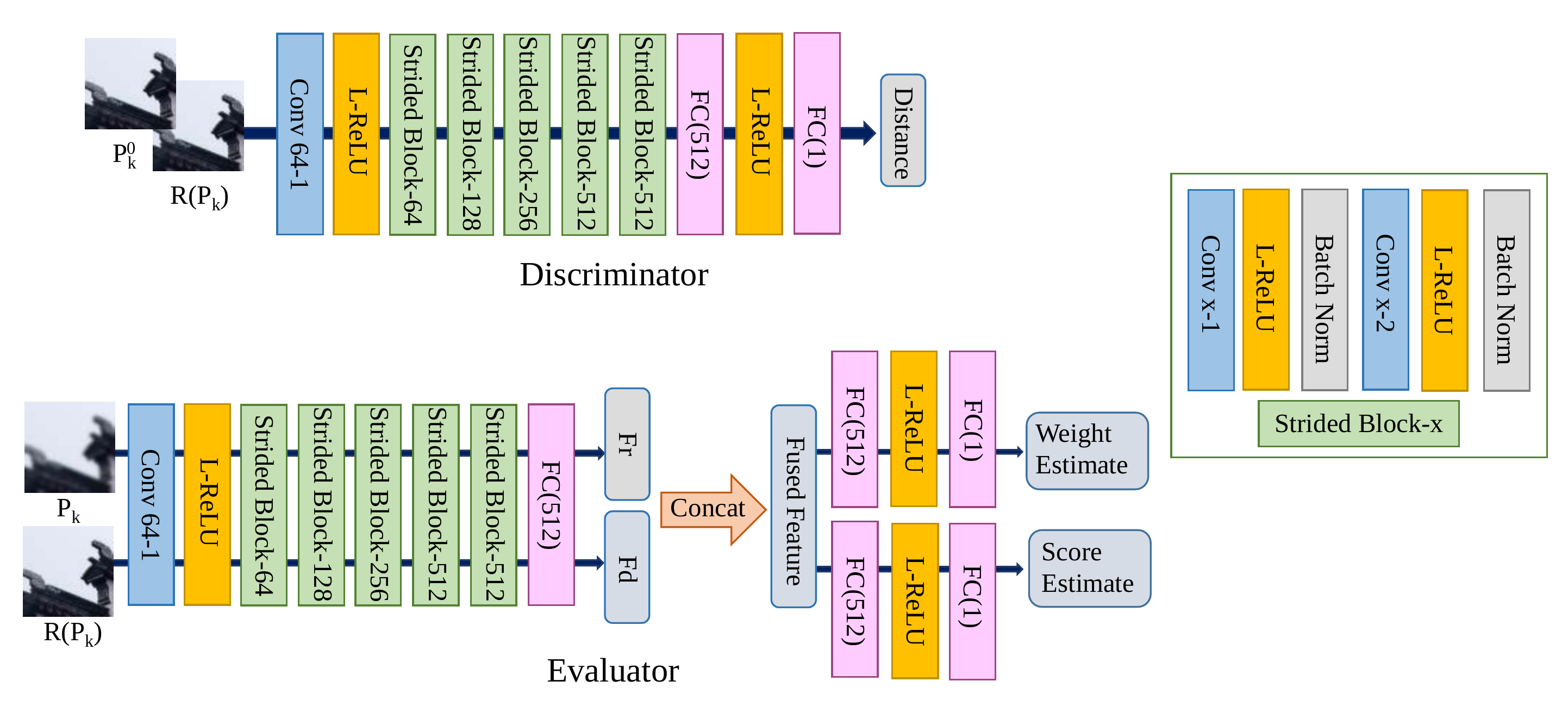}
\caption{Overview of the detailed architecture of RAN. From up to down are the structures of restorator, discriminator and evaluator. ``Conv $m$-$n$'' denotes a convolution layer with $m$ feature maps and 3$\times$3 filters with stride $n$. ``L-ReLU'' denotes the Leaky ReLU.}
\label{Model}
\end{figure*}

Inspired by how HVS works, we hereby define Gain of Restoration (GoR) and assess perceptual quality using it. Given a distorted image, GoR is the perceptual distance between the distorted image and its restored version. In this paper, we propose restorative adversarial nets (RAN), an NR-IQA model which composes of a restorator, a discriminator and an evaluator. The proposed model is shown in Fig.~\ref{Model}. Different from the original generative adversarial nets (GAN)~\cite{goodfellow2014generative}, the restorator takes a distorted image patch as input instead of Gaussian noise and it is trained to restore the input. The discriminator aims to discern whether the input belongs to a restored version or is originally distortion-free. The evaluator takes restored and distorted patches as input and outputs perceptually quantified score by fitting and utilizing the monotonicity of GoR. The proposed model adopts perceptual loss~\cite{johnson2016perceptual,ledig2016photo} to achieve perceptually friendly results in restoration and Wasserstein distance~\cite{arjovsky2017wasserstein} to stabilize adversarial training.

The remainder of this paper is organized as follows. In section \uppercase\expandafter{\romannumeral2}, we introduce related work in the field of NR-IQA and GAN. We detail the structure of RAN in section \uppercase\expandafter{\romannumeral3}. Experimental results and related discussions are presented in section \uppercase\expandafter{\romannumeral4}. In section \uppercase\expandafter{\romannumeral5}, we conclude the paper.

\section{Related Work}
\subsection{No-Reference Image Quality Assessment}

Given a no-reference setting, \cite{moorthy2011blind} propose DIVINE framework, which deploys summary statistics and performs a regression after identification of distortion type. BRISQUE~\cite{saad2012blind} and BLIINDS-II~\cite{mittal2012no} utilize natural scene statistics (NSS) to model perceptual quality. CORNIA~\cite{ye2012unsupervised}, on the other hand, constructs and presents a small yet accurate codebook to look up.~\cite{bosse2016deep,kang2015simultaneous,kang2014convolutional} adopt deep neural network to extract features from the raw input and perform regression to estimate perceptual scores. The above NR-IQA methods can be summarized as feature extraction and regression only based on distorted images. However, according to the free-energy theory~\cite{friston2006free,friston2010free}, HVS tends to restore the distorted image first before quality assessment. Despite building NR-IQA model based on the free-energy theory, \cite{zhai2012psychovisual} and~\cite{gu2015using} restore the distorted image with a linear autoregressive model, which is not capable of producing a satisfactory result when the input suffers high-level distortion and therefore may not be consistent with HVS.

\subsection{Generative Adversarial Nets}
Proposed by~\cite{goodfellow2014generative}, generative adversarial nets (GAN) have achieved impressive success in style transfer~\cite{zhu2017unpaired}, image super resolution~\cite{ledig2016photo}, and representation learning~\cite{radford2015unsupervised}. The key idea of GAN is to train a generator and a discriminator simultaneously, in which the generator takes noise as input and tries to fool the discriminator by generating indistinguishable samples. However, it suffers vanishing gradient and mode collapse problems when the discriminator is well-trained~\cite{salimans2016improved}. To address the problems, \cite{arjovsky2017wasserstein} propose Wasserstein GAN (WGAN), which greatly reduces instability of training. In this paper, stability is critical since the inputs are images with various kinds and levels of distortion, so we adopt WGAN framework to stabilize adversarial training.

\section{Proposed Model and Learning}
In NR-IQA, given a distorted image ${I_d}$, the aim is to learn a mapping $f:{I_d}\to{s}$, in which ${s}\in{\mathbb{R^{+}}}$ denotes the quality estimate of ${I_d}$ and should be consistent with the result of human visual system (HVS). As  shown in Fig.~\ref{Model}, the proposed model consists of three parts: a restorator $R_\theta$, a discriminator $D_\phi$ and an evaluator $E_\omega$. They are realized by neural networks parametrized by $\theta$, $\phi$, $\omega$ respectively. It works as follows. We first sample non-overlapping patches ${\psi=\{P_0,P_1,\ldots,P_n\}}$ from the given distorted image ${I_d}$. For each patch ${P_k}\in{\psi}$, $R_\theta$ takes it as input and tries to restore ${P_k}$ into corresponding distortion-free pristine patch ${P_{k}^{0}}$, $D_\phi$ distinguishes restored $R_\theta({P_k})$ from pristine ${P_{k}^{0}}$. $E_\omega$ takes both ${P_k}$ and $R_\theta({P_k})$ as inputs and outputs $s_k$ and $w_k$, denoting a quality estimate and weight estimate of the patch ${P_k}$. Eventually, the perceptual quality $Q({I_d})$ is estimated by weighted sum of score predictions of all patches. 
\begin{equation}Q(I_d)=\frac{\sum_{k=1}^n s_k w_k}{\sum_{k=1}^n w_k}\end{equation}

\begin{figure*}
\centering
\subfigure{
\includegraphics[width=0.24\textwidth]{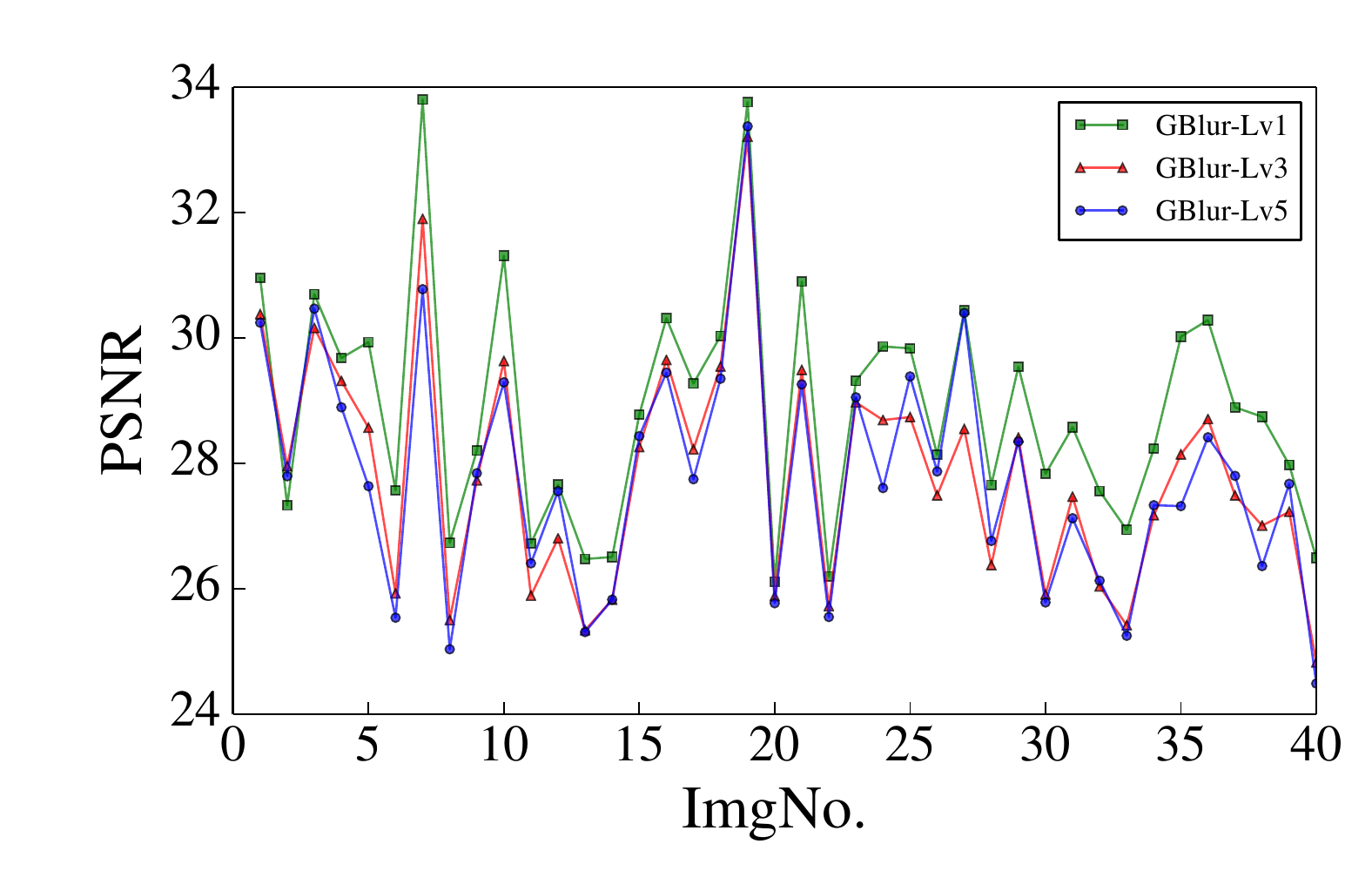}}
\subfigure{
\includegraphics[width=0.24\textwidth]{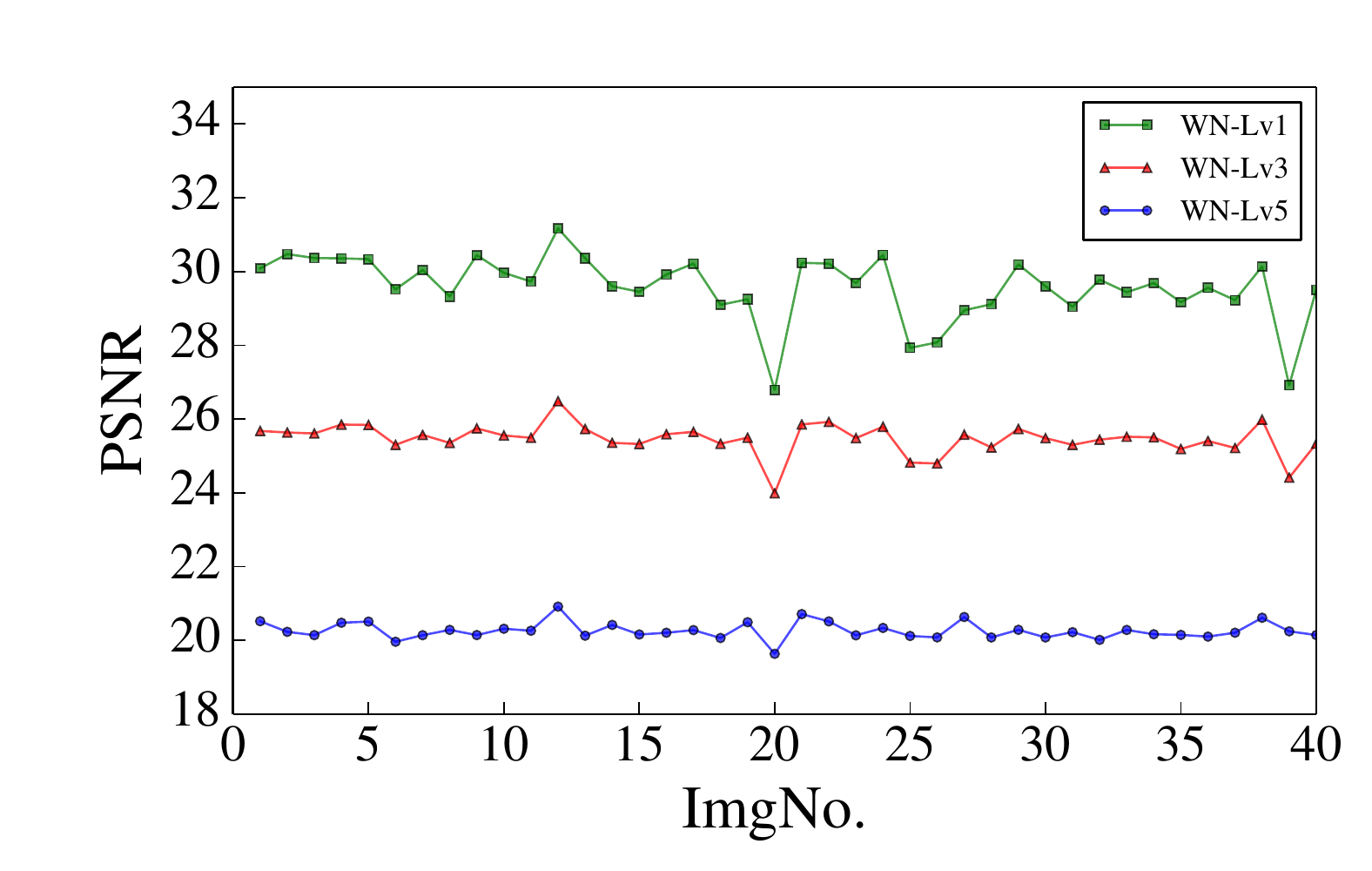}}
\subfigure{
\includegraphics[width=0.24\textwidth]{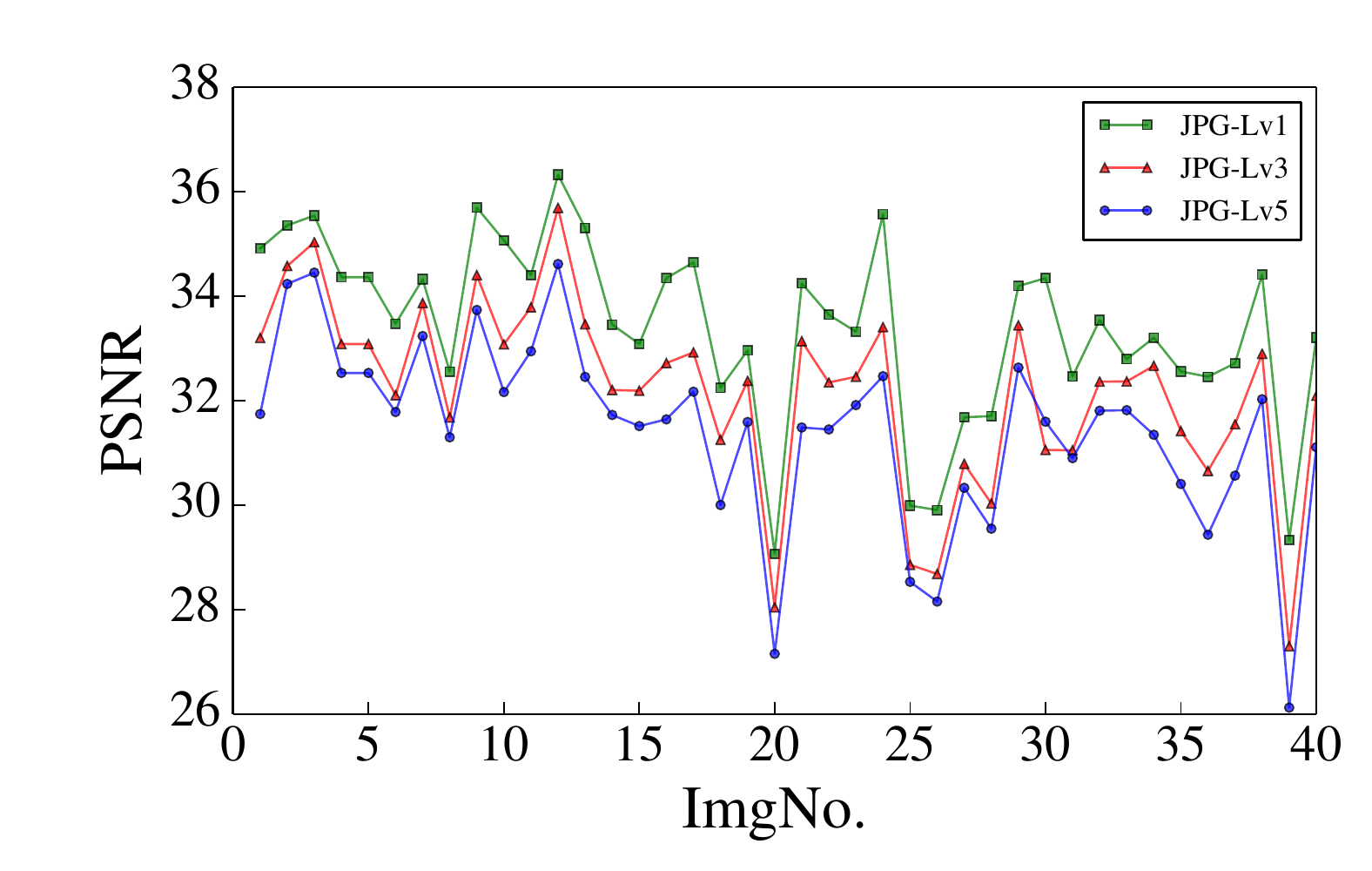}}
\subfigure{
\includegraphics[width=0.24\textwidth]{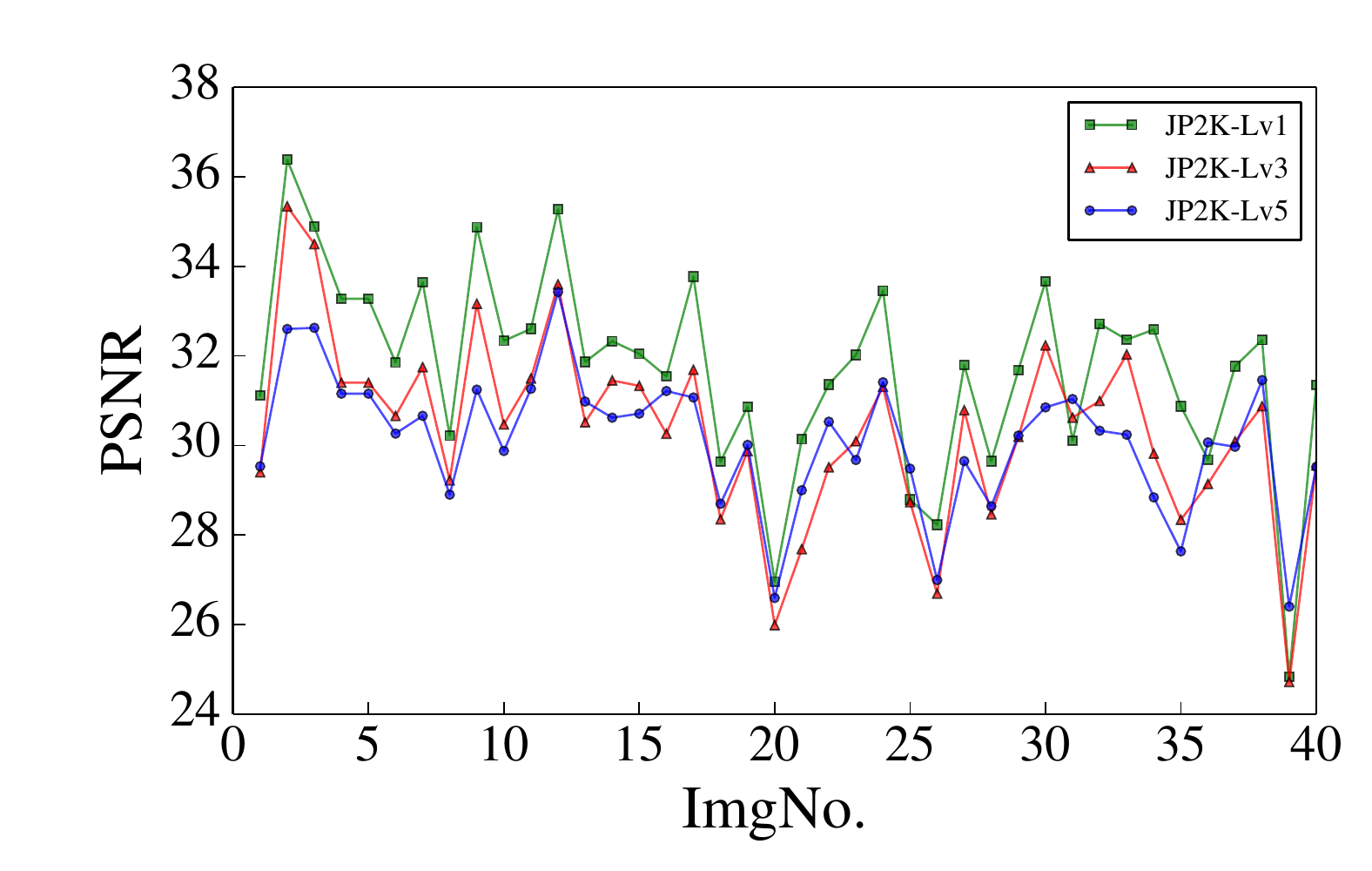}}
\subfigure{
\includegraphics[width=0.24\textwidth]{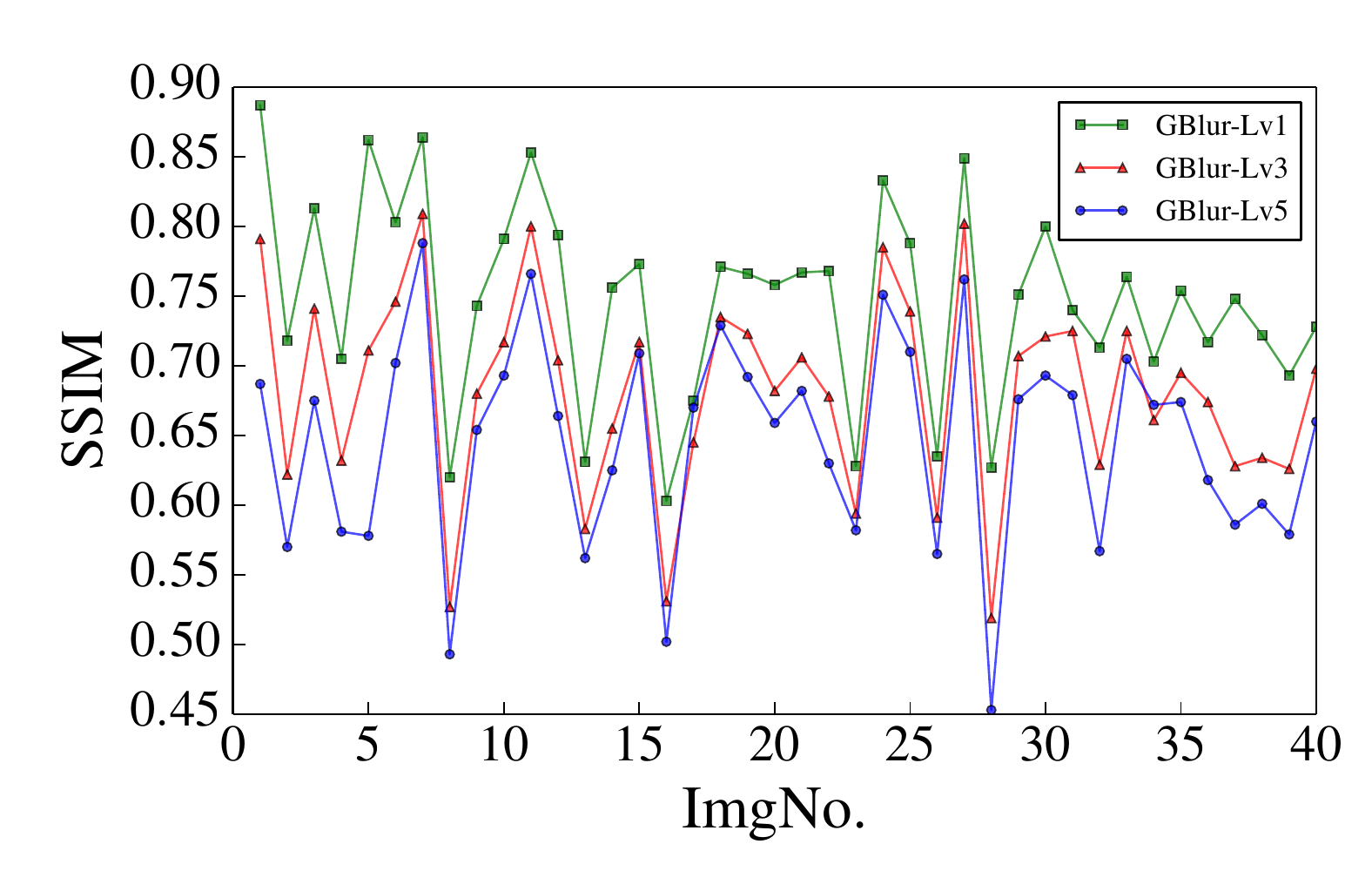}}
\subfigure{
\includegraphics[width=0.24\textwidth]{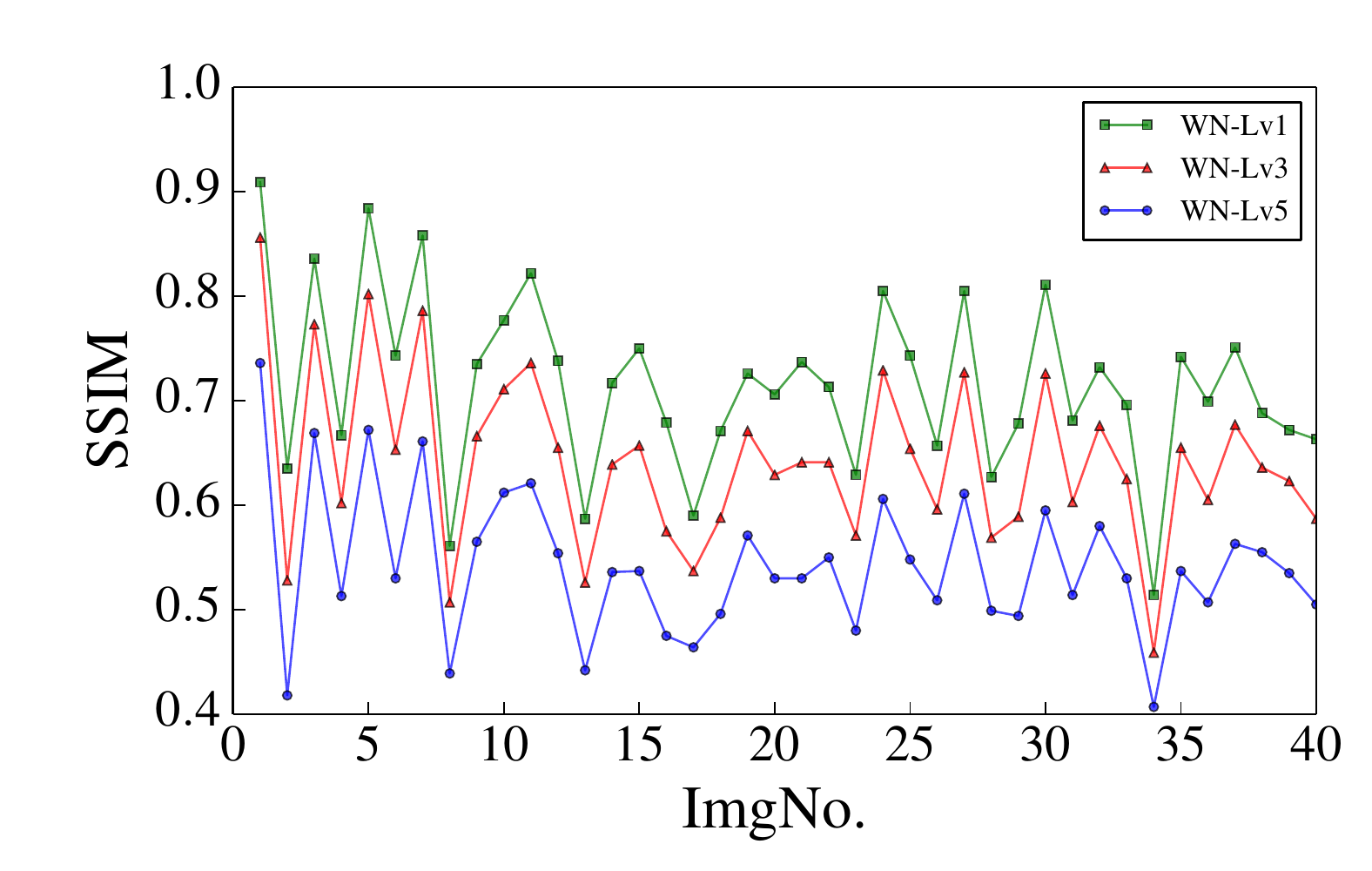}}
\subfigure{
\includegraphics[width=0.24\textwidth]{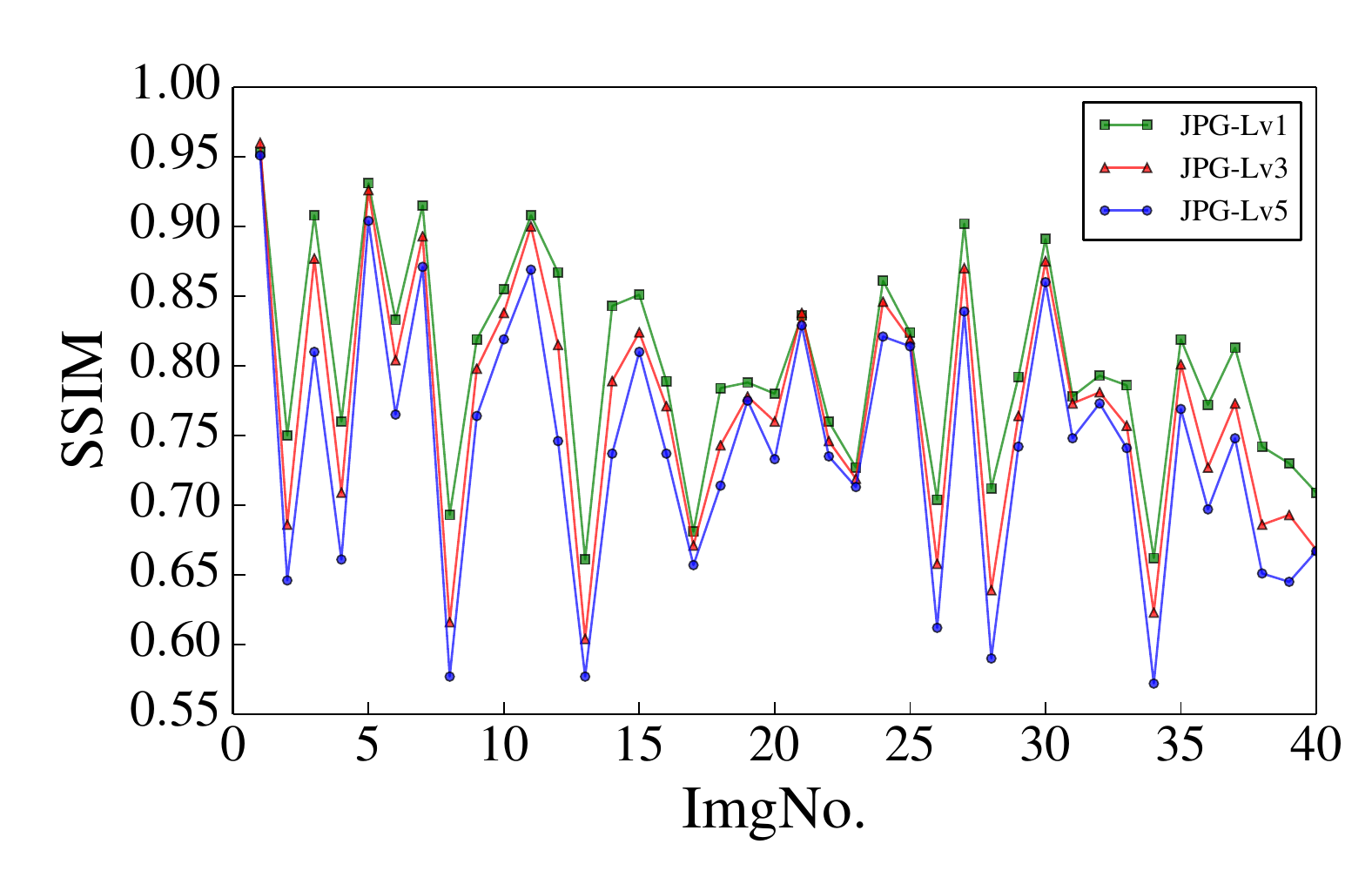}}
\subfigure{
\includegraphics[width=0.24\textwidth]{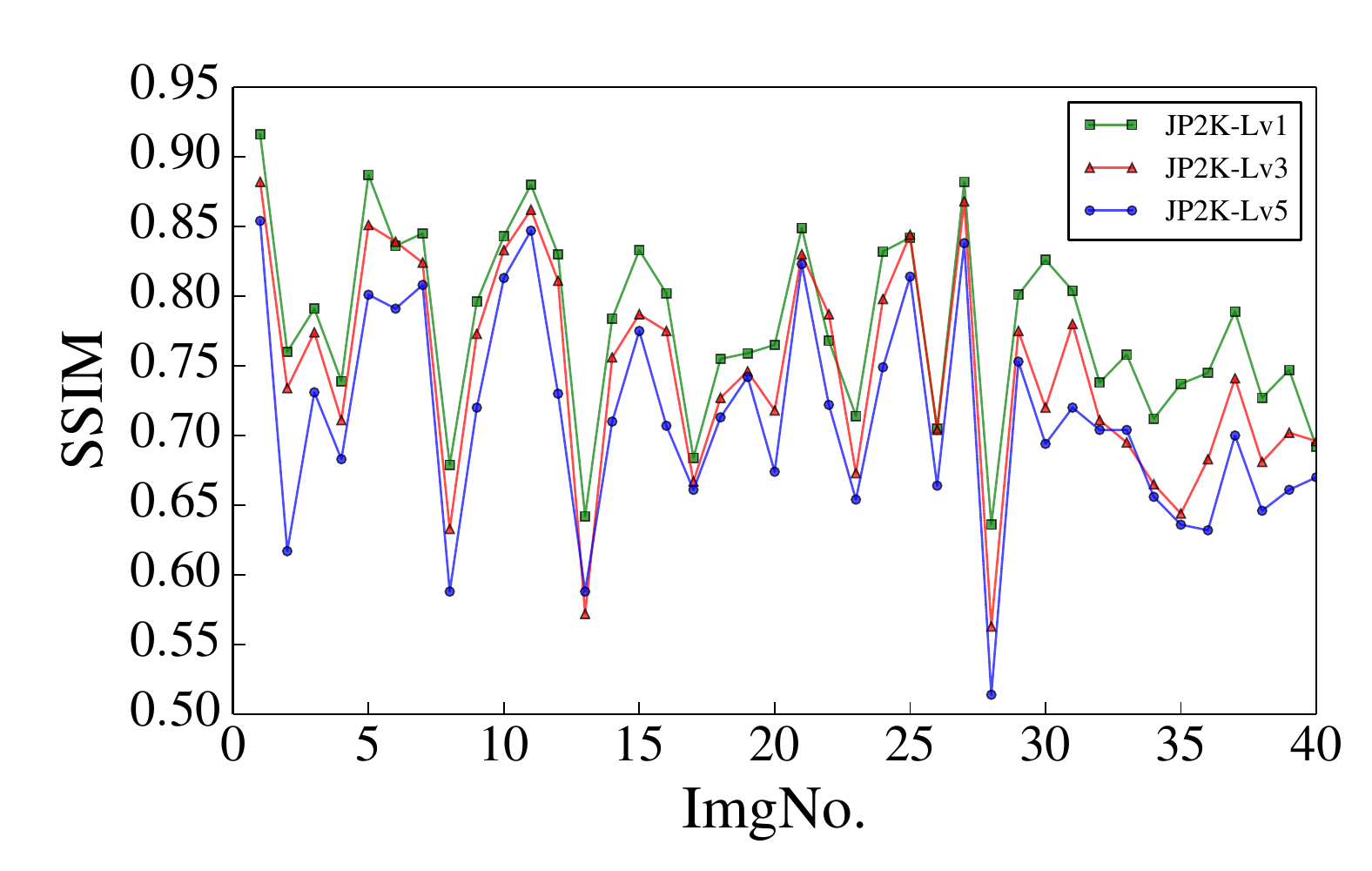}}
\caption{Monotonicity of GoR. We randomly select 40 pristine images in Waterloo Exploration~\cite{ma2017waterloo} and their distorted version with 4 types and 3 levels of distortions. The distorted images are restored and we measure the distance between the distorted and restored images. The vertical axis denotes perceptual distance, measured by PSNR or SSIM~\cite{wang2004image}, between distorted image and restored version. The four charts on the first row are PSNR-measured, the four below are SSIM-measured. From left to right, the distortion type is Gaussian Blur, White Noise, JPEG Compression and JP2K Compression. Green, red, blue represent distortion level of 1, 3, 5 respectively. Severer distortion results in larger distance.}
\label{Mono}
\end{figure*}

\subsection{Gain of Restoration}
We define Gain of Restoration (GoR) as the perceptual distance between restored image and distorted image in the proposed NR-IQA model. We observe that GoR is monotone with respect to distortion level given a determined restorator. 
As shown in Fig.~\ref{Mono}, we try several functions to measure the perceptual distance, such as PSNR and SSIM~\cite{wang2004image}, after the restorator is trained. PSNR and SSIM are two widely-used metrics to measure the difference between two given images. We randomly picked 40 pristine images and their corresponding distorted version from Waterloo Exploration~\cite{ma2017waterloo}, a large-scale IQA database. We select all 4 distortion types: Gaussian Blur, White Noise, JPEG Compression and JP2K Compression. Waterloo Exploration generates distortion of 5 levels for each type, however, to justify GoR monotonicity, we choose 3 distortion levels: level 1, 3 and 5. After 480 distorted images are restored, we calculate PSNR and SSIM between distorted images and their restored version. The four charts in the first row show PSNR results, representing Gaussian Blur, White Noise, JPEG Compression and JP2K Compression from left to right. The four charts in the second row are SSIM results. We observe that perceptual similarity, quantified by SSIM and PSNR, between the restored image and the distorted image is low when the image has severe distortion, which means restoration is a huge improvement and GoR is high, and vice versa.

Although 
restoration is a poorly-defined problem because the restorator is trained to restore the inputs with various types and levels of distortion to a single distortion-free image,  we aim to maximize GoR of every input, and assess perceptual quality based on GoR, instead of focusing on how close the restored image is to the pristine image.

\subsection{Restorative Adversarial Nets}
The restorator $R_\theta$ takes in distorted patch $P_k$ and restores the input ${P_k}$ towards the pristine patch ${P_{k}^{0}}$. Motivated by~\cite{Gross2016residual,ledig2016photo}, we adopt the layout of a deep ResNet~\cite{he2016deep} for restorator $R_\theta$. Residual structure is critical in our proposed model. First, according to~\cite{he2016deep}, residual connections make the identity function easier to train, which is crucial in NR-IQA setting, because it is desired to restore images with low level of distortion as well as ones with high level of distortion. Second, ResNet structure is similar to the way HVS restores: adding details and textures onto distorted input chronologically. Gradually adding residual information simulates HVS model and greatly reduces potential color shifting problems in stacked convolution-ReLU networks. Specifically, each residual block has identical layout which has two convolutions with 3$\times$3 filters and 64 feature maps. We also adopt batch normalization~\cite{ioffe2015batch} to avoid undesired initialization, branches are added element-wisely to accumulate residue. After 10 identical residual blocks, we achieve the restored image with 3 channels by a convolution.

As for the discriminator $D_\phi$, which aims to distinguish restored image patch $R_\theta({P_k})$ from ${P_{k}^{0}}$, we propose a VGG-based~\cite{simonyan2014very} discriminator due to its successful application in several computer vision tasks. The discriminator has a similar layout of 13-layer VGG. Instead of pooling, we follow the advice of~\cite{radford2015unsupervised} and use strided convolution layers to downsample. Furthermore, we also adopt batch normalization~\cite{ioffe2015batch} after each Leaky ReLU activation layer. After extracting 512 feature maps, we use two fully connected layers to approximate discriminator objectives.


The loss function for RAN consists of a perceptual loss (realized by a pretrained VGG network) and an adversarial loss (realized by approximating Wasserstein distance), which is shown in Eq.~\ref{equ:ranloss}. 

\begin{equation}
\label{equ:ranloss}
\mathcal{L}_{RAN} = \min_\theta \max_\phi (\mathcal{L}^{PER}_{RAN} + \mathcal{L}^{ADV}_{RAN})
\end{equation}

\paragraph{Perceptual loss}
Previous computer vision research relies on pixel-wise loss, such as mean squared error (MSE) to measure the distance between two given images. However, according to~\cite{wang2009mean}, pixel-wise loss fails to capture perceptual representations, such as texture details, as humans do. Lower MSE does not necessarily reflect better perceptual similarity. In this regard, we adopt perceptual loss~\cite{johnson2016perceptual} as cost function for the restorator. Perceptual loss is defined on a VGG19 network which is pretrained on IMAGENET~\cite{deng2009imagenet}. Instead of measuring pixel-by-pixel difference, perceptual loss quantifies perceptual difference based on various levels of feature representations extracted by the convolution layers in the pretrained network. The inputs are mapped into the feature space by non-linear differentiable functions, which in our setting, are the convolution layers from the pretrained model. In detail, the pretrained VGG19 has five maxpool layers and we select the last convolution layers before the corresponding five maxpool activations. Perceptual loss function is defined as the Euclidean distance of the five convolution layers, the $i^{th}$ convolution $\Omega_i$ is of size $W_{i}\times{H_{i}}\times{C_{i}}$.
\begin{equation}\label{equ:rlossper}
\mathcal{L}^{PER}_{RAN}=\sum_{i}{\frac{1}{W_{i}H_{i}C_{i}}\parallel{\Omega_{i}{(P_k^{0})}-\Omega_{i}{(R_\theta(P_k))}}\parallel_{2}^{2}}\end{equation}

\paragraph{Wasserstein GAN} While traditional GANs suffer vanishing gradient and mode corruption problems, \cite{arjovsky2017wasserstein} propose  Wasserstein GAN (WGAN), which claims to solve the problems with theoretical analysis.

Instead of minimizing the Jensen-Shannon divergence, which leads to vanishing gradient when the discriminator reaches optima, a loss function which fits the Earth-Mover distance is defined as the cost between generated distribution and real distribution. Unlike its traditional counterpart which does 0-1 classification, the discriminator in WGAN model solves a regression problem. According to~\cite{arjovsky2017wasserstein}, when the discriminator is K-Lipschitz, the Earth-Mover distance suffers less gradient saturation and mode corruption problems even when the support of the generated distribution and the real distribution does not have a non-negligible intersection.
In our model, WGAN is crucial because the inputs of the restorator are images with various types and levels of distortions, stability is desired. As shown in Fig.~\ref{Model}, we remove the sigmoid layer in the discriminator. Furthermore, to enforce K-Lipschitz on discriminator, we clip weights between [-0.05, 0.05] after each update. Lastly, we use RMSProp~\cite{tieleman2012lecture} optimization to further avoid instability.
\begin{equation}\label{equ:rlossadv}
\mathcal{L}^{ADV}_{RAN} = D_\phi(P_k^0) - D_\phi(R_\theta(P_k))\end{equation}

\subsection{Evaluator}
After RAN restoration, we further train an evaluator $E_\omega$ to quantify visual score based on GoR monotonicity. As shown in Fig.~\ref{Model}, still patch-wisely, evaluator takes the restored patch and the distorted patch as input. Similar to the discriminator's function, we adopt the same layout to first extract feature representations. Second, feature vectors extracted from distorted patch and restored patch are fused and concatenated into a 1024-dimensional vector. Since the distortion is not evenly distributed, an average of scores of all patches does not necessarily reflect the global perceptual quality of the full image. Thus, in the last stage, the fused feature vector is fed into 2 branches which calculate the perceptual score $s_k$ and weight $w_k$ respectively. The whole model collects the weighted sum of all patches, which is the final quality estimate of distorted image $I_d$.

The loss function of evaluator is shown in Eq.~\ref{equ:eloss1} and Eq.~\ref{equ:eloss2}. The evaluator is trained for two steps, which will be detailed in the experiments. When the score and weight labels for every image patch are available, we minimize the mean average error between the prediction and label for score and weight. 

\begin{equation}\label{equ:eloss1}
\mathcal{L}_{Eva}=\sum_{k=1}^{n}{\left|s_{k}-s_{k}^{0}\right|} + \sum_{k=1}^{n}{\left|w_{k}-w_{k}^{0}\right|}
\end{equation}

However, when we only have access to the score of the whole image, the mean average error between ground truth and weighted prediction is minimized by each update.

\begin{equation}\label{equ:eloss2}
\mathcal{L}_{Eva}=\left|\frac{\sum_{k=1}^{n}{s_{k}w_{k}}}{\sum_{k}{w_k}}-s\right|\end{equation}

\noindent where $s_{k}$ and $w_{k}$ denote the score and weight prediction for image patch $P_k$. $s_{k}^{0}$ and $w_{k}^{0}$ denote the ground truth score and weight for patch $P_k$. $s$ represents the ground truth score for the whole image $I_d$.
\section{Experimental Results}
In this section, we conduct several experiments to test the performance of RAN on several datasets. We pretrain RAN on Waterloo Exploration~\cite{ma2017waterloo}, perform cross validation on TID2013~\cite{ponomarenko2015image} and LIVE~\cite{sheikh2005live}. Furthermore, we conduct ablation experiments to test and quantify the performance gain and the necessity of each component and technique.
\subsection{Protocol}
\paragraph{TID2013} TID2013 is an extended version of TID2008 \cite{ponomarenko2009tid2008}, which consists of 25 distortion-free reference images and 3000 distorted images created from references at five levels. There are 24 different distortion types, ranging from additive Gaussian noise to sparse sampling reconstruction. Its wide range makes it one of the most comprehensive IQA databases. Mean Opinion Scores (MOS) are provided for every image, which is in the range [0, 9] and higher MOS means higher perceptual quality.

\paragraph{LIVE} LIVE comprises 29 reference images and 779 distorted samples with 5 distortion types: JPEG Compression, JP2K Compression, White Noise, Gaussian Blur and Fast Fading. Every image is annotated with Differential Mean Opinion Scores (DMOS), which is in the range [0, 100]. Lower DMOS means higher perceptual quality. DMOS value of zero indicates the image is distortion-free.

\paragraph{Waterloo Exploration} Waterloo Exploration is a large-scale IQA database, which contains 4744 pristine natural images and 94880 distorted images generated by MATLAB scripts with four distortion types and five levels. Compared to TID2013 and LIVE, Waterloo Exploration has a great diversity of image content. The four types: JPEG Compression, JP2K Compression, Gaussian Blur and White Noise, are also considered the most common distortion types and are covered both in TID2013 and LIVE. Instead of annotating distorted data with subjective mean opinion score (MOS), which is impractical for such a large database, Waterloo Exploration claims to preset MATLAB parameters which cover a wide range of subjective quality scale.

We adopt two measures to evaluate the performance of RAN: Spearman rank order correlation coefficient (SROCC) and Pearson linear correlation coefficient (PLCC). SROCC measures the prediction monotonicity while PLCC takes relative distance into consideration and thus a non-linear regression is performed.

\begin{table*}[tbp]
\begin{center}
\begin{tabular}{c|l|cc|cc}
\hline
&\multirow{2}*{IQA methods}&\multicolumn{2}{c}{TID2013}&\multicolumn{2}{c}{LIVE}\\
\cline{3-6}
& &SROCC &PLCC &SROCC &PLCC\\
\hline
\multirow{5}*{FR-IQA}&PSNR &0.889 &0.847 &0.880 &0.805\\
&SSIM~\cite{wang2004image} &0.856 &0.867 &0.918 &0.780\\
&FSIM~\cite{zhang2011fsim} &$\bm{0.963}$ &0.932 &$\bm{0.952}$ &0.822\\
&FSIM$_{C}$~\cite{zhang2011fsim} &$\bm{0.963}$ &0.935 &0.951 &0.816\\
&VSI~\cite{zhang2014vsi} &0.947 &$\bm{0.939}$ &0.936 &$\bm{0.853}$\\
\hline
\multirow{7}*{NR-IQA}&DIIVINE~\cite{moorthy2011blind} &0.855 &0.851 &0.885 &0.853\\
&BLIINDS-II~\cite{mittal2012no} &0.877 &0.841 &0.931 &0.930\\
&BRISQUE~\cite{saad2012blind} &0.922 &0.917 &0.940 &0.911\\
&CNN~\cite{kang2014convolutional} &0.903 &0.917 &0.913 &0.888\\
&CNN++~\cite{kang2015simultaneous} &0.843 &0.804 &0.928 &0.897\\
&DNN~\cite{bosse2016deep} &0.933 &0.909 &0.960 &$\bm{0.972}$\\
&RAN (proposed) &$\bm{0.948}$ &$\bm{0.937}$ &$\bm{0.972}$ &0.968\\
\hline
\end{tabular}
\caption{Cross Validation on TID2013 and LIVE}
\end{center}
\label{Result}
\end{table*}

\subsection{Training Details}
All the training is performed on an NVIDIA Tesla K40 GPU. The training consists of two steps: (1) pretrain on the Waterloo Exploration, (2) finetune on the TID2013 and LIVE. We crop every image into 64$\times$64 non-overlapping patches. The implementation is on tensorflow~\cite{abadi2016tensorflow}.

In the pretrain step, we first train the restorator based on the labels (pristine patches) to avoid unwanted minima using Adam optimizer~\cite{kingma2014adam} at a learning rate of $10^{-4}$ for $300,000$ iterations. Then we train the restorator and the discriminator together using RMSProp~\cite{tieleman2012lecture} at a learning rate of $10^{-4}$ for $300,000$ iterations and a lower learning rate of $10^{-5}$ for another $300,000$ iterations. In each iteration, we train the discriminator 5 times and the restorator once. Then we freeze the weights of restorator and discriminator, and pretrain the evaluator using Adam at a learning rate of $10^{-4}$ for $300,000$ iterations. Note that all the above training uses $4744$ pristine images and $94,880$ distorted images from Waterloo Exploration. Considering that perceptual scores are not available in Waterloo Exploration, we label a score and weight for every distorted image patch by performing FSIM~\cite{zhang2011fsim}, which is one of the state-of-the-art FR-IQA metrics, on each patch of Waterloo Exploration. These scores and weights serve as the ground truth label in the pretrain of the evaluator. Note that the generated FSIM labels are only used during the pretrain of the evaluator.

In the finetune step, we also freeze the weights of the restorator and discriminator, only train the evaluator using the same optimizer and learning rate on different datasets, which will be elucidated in the next section.


\subsection{Cross Validation on TID2013 and LIVE}
After pretrained on Waterloo Exploration, the evaluator is finetuned on TID2013 and LIVE to perform cross validation respectively. Since Waterloo Exploration only contains 4 distortion types, we finetune and test RAN on them: Gaussian Blur, White Noise, JPEG and JP2K.

On TID2013, we randomly pick 60\% as the training set, 20\% as the validation set and the left 20\% as the test set. We finetune the evaluator for $20,000$ iterations. The result is displayed in Table 1. The best results of both NR-IQA metrics and FR-IQA metrics are in bold. The proposed model achieves as good a result as FSIM, on which RAN relys during pretrain. RAN outperforms state-of-the-art NR-IQA metrics in SROCC and PLCC in TID2013.

We also perform two-sided t-test on average SROCC and PLCC between the proposed model and all the other mentioned NR-IQA models. The null hypothesis is that the two models have equal SROCC at the 95\% confidence level. An alternative hypothesis is that our model has higher/less correlation results. We also make a similar hypothesis on PLCC. We randomly split the TID2013, train, test for 15 times and achieve the results for all the models. Our model is statistically significant than all the other NR-IQA models on both SROCC and PLCC.

To further test the robustness and generalization of RAN, we test our model on LIVE, where we split the dataset into 6:2:2 for train, validation and test respectively. LIVE has fewer images compared to TID2013, we finetune the evaluator for $15,000$ iterations. As shown in Table 1, RAN outperforms the state-of-the-art metrics, which shows its robustness and generalization across datasets.

\subsection{Ablation Experiments}
To demonstrate that the adopted techniques are critical for the performance, we conduct several ablation experiments on TID2013, in which we remove perceptual loss, Wasserstein distance, discriminator, weighted strategy and test the performance of the remaining framework. When we remove perceptual loss (RAN$\backslash$PER), we use L2 loss instead during the training of restorator. When we remove Wasserstein (RAN$\backslash$WAS), we adopt the original logarithm-based adversarial loss function. When we remove the discriminator (RAN$\backslash$DIS), the only component left is the restorator, which is trained only based on the perceptual loss objective. Without weighted strategy (RAN$\backslash$WEI), we simply estimate the quality of an image as the average of patch scores. As shown in Table 2, removal of the discriminator causes significant performance decline because restorator is not propelled by discriminator. The weighted strategy is also critical for the evaluator since distortion is not evenly distributed on test images, a naive average strategy will not give reasonable result when the image has severe distortion in a small area. Not using perceptual loss affects performance because without an HVS-consistent restorator, GoR monotonicity will not be notable and it is hard for the evaluator to assess perceptual quality. RAN without Wasserstein distance performs slightly worse than the proposed model.

\begin{table}[tbp]
\begin{center}
\begin{tabular}{l|cc}
\hline
\multirow{2}*{Ablation}&\multicolumn{2}{c}{TID2013}\\
\cline{2-3}
&SROCC &PLCC\\
\hline
RAN (proposed) &$\bm{0.948}$ &$\bm{0.937}$\\
RAN$\backslash$PER &0.893 &0.879\\
RAN$\backslash$WAS &0.936 &0.906\\
RAN$\backslash$DIS &0.854 &0.856\\
RAN$\backslash$WEI &0.857 &0.884\\
\hline
\end{tabular}
\caption{Ablation Experiment on TID2013}
\end{center}
\label{Result}
\end{table}

\section{Conclusion}
This paper presents RAN, a GAN-based no-reference image quality assessment model. Consistent with the human visual system, RAN restores input distorted image, extracts features both from distorted image and restored image based on GoR and evaluates perceptual quality by comparing them. Experimental results on standard IQA database have shown its superiority over state-of-the-art IQA methods and its generalization capacity.

\section{Acknowledgement}
We would like to express our thanks for support from the following research grants: 973-2015CB351800, NSFC-61527804, NSFC-61625201, NSFC-61421062 and NSFC-61210005.

\bibliography{aaai}
\bibliographystyle{aaai}
\end{document}